%% file: goat.tex
\pdfoutput=1

\documentclass{article}
\usepackage{goat}
\usepackage[export]{adjustbox}
\usepackage{multirow}
\usepackage{amsmath}
\usepackage{advdate}
\usepackage{mathtools}
\usepackage{subcaption}
\newcommand{\comment}[1]{}
\usepackage{float}
\usepackage{placeins}
\usepackage{natbib}  

\usepackage{blindtext}
\input{math_commands.tex}

\usepackage{amsthm}
\usepackage[scientific-notation=true]{siunitx}
\usepackage{amsthm}
\usepackage{float}
\usepackage{placeins}
\theoremstyle{definition}
\newtheorem{definition}{Definition}[section]

\newtheorem{proposition}{Proposition}[section]

\usepackage{xcolor}

\usepackage{times}
\usepackage{soul}
\usepackage[hyphens]{url}
\usepackage[hidelinks,bookmarks=false]{hyperref}
\usepackage{nccmath}    

\usepackage[utf8]{inputenc}
\usepackage{caption}
\usepackage{graphicx}
\usepackage{amsmath}
\usepackage{amsthm}
\usepackage{booktabs}
\usepackage{algorithm}
\usepackage{algorithmic}
\usepackage{amsmath, amsthm, amssymb}
\usepackage{booktabs}

\makeatletter
\newcommand{\ALOOP}[1]{\ALC@it\algorithmicloop\ #1%
  \begin{ALC@loop}}
\newcommand{\ENDALOOP}{\end{ALC@loop}\ALC@it\algorithmicendloop}

\newcommand{\algorithmicbreak}{\textbf{break}}
\newcommand{\BREAK}{\STATE \algorithmicbreak}
\makeatother

\newcommand\numberthis{\addtocounter{equation}{1}\tag{\theequation}}
\newcommand{\eqnref}[1]{(\ref{#1})}

\newcommand{\neighborhood}[1]{\gN({#1})}
\newcommand{\closedneighborhood}[1]{\overline{\gN}({#1})}
\newcommand{\cnsize}{P}
\newcommand{\layersuperscript}{l}
\newcommand{\gso}{\mS}
\newcommand{\recfn}{\rho}
\newcommand{\feedfn}{\phi}

\DeclareMathOperator*{\concat}{\scalebox{1}[1.5]{$\parallel$}}

\newtheorem{theorem}{Theorem}

\title{Graph Ordering Attention Networks}

\author{
Michail Chatzianastasis$^1$\footnote{Contact Author}\and
Johannes F. Lutzeyer$^1$\and
George Dasoulas$^1$\And
Michalis Vazirgiannis$^1$\\
\affiliations
$^1$ DaSciM, LIX, École Polytechnique, Institut Polytechnique de Paris, France.\\
\emails
\{michail.chatzianastasis, ~johannes.lutzeyer\}@polytechnique.edu, \\
george.dasoulas1@gmail.com, \quad
mvazirg@lix.polytechnique.fr
}

\begin{document}

\maketitle

\begin{abstract}
Graph Neural Networks (GNNs) have been successfully used in many problems involving graph-structured data, achieving state-of-the-art performance. 
GNNs typically employ a message-passing scheme, in which every node aggregates information from its neighbors using a permutation-invariant aggregation function.
Standard well-examined choices such as the mean or sum aggregation functions have limited capabilities, as they are not able to capture interactions among neighbors. 
In this work, we formalize these interactions using an information-theoretic framework that notably includes synergistic information. 
Driven by this definition, we introduce the Graph Ordering Attention (GOAT) layer, a novel GNN component that captures interactions between nodes in a neighborhood. 
This is achieved by learning local node orderings via an attention mechanism and processing the ordered representations using a recurrent neural network aggregator. 
This design allows us to make use of a permutation-sensitive aggregator while maintaining the permutation-equivariance of the proposed GOAT layer. 
The GOAT model demonstrates its increased performance in modeling graph metrics that capture complex information, such as the betweenness centrality and the effective size of a node. In practical use-cases, its superior modeling capability is confirmed through its success in several real-world node classification benchmarks. 
\end{abstract}

\section{Introduction}

Graph Neural Networks (GNNs) achieve remarkable success in machine learning problems on graphs \citep{4700287,kipf2017semisupervised,bronstein2021geometric}. In these problems, data arises in the structure of attributed graphs, where in addition to the node and edge sets defining a graph, a set of feature vectors containing data on each node is present. 
The majority of GNNs learn node representations using a message-passing scheme \citep{gilmer2017neural}. In such message passing neural networks (MPNN) each node iteratively aggregates the feature vectors or hidden representations of its neighbors to update its own hidden representation. Since no specific node ordering exists, the aggregator has to be a permutation-invariant function \citep{xu2019powerful}. 

Although MPNNs have achieved great results, they have severe limitations. Their permutation-invariant aggregators treat neighboring nodes as a set and process them individually, omitting potential interactions between the large number of subsets that the neighboring nodes can form. Therefore, current MPNNs cannot observe the entire structure of neighborhoods in a graph \citep{pei2020geomgcn} and cannot capture all \textit{synergistic interactions} between neighbors \citep{murphy2019janossy,wagstaff2021universal}.

The concept of synergy is important in many scientific fields and is central to our discussion here.  It expresses the fact that some source variables are more informative when observed together instead of independently.
For example in neuroscience, synergy is observed when the target variable corresponds to a stimulus and the source variables are the responses of different neurons \citep{10.3389/fncom.2013.00051}. 
Synergistic information is often
presented in biological cells, 
where extra information is provided by patterns of coincident spikes from several neurons \citep{article_synergy}.
In gene-gene interactions, synergy is present when the contribution of two mutations to the phenotype of a double mutant is larger than the expected additive effects of the individual mutations \citep{PEREZPEREZ2009368}. 
We believe the consideration of synergistic information to have great potential in the GNN literature. 

In this paper, to better understand interactions between nodes, we introduce the Partial Information Decomposition (PID) framework \citep{pid} to the graph learning context. 
We decompose the information that neighborhood nodes have about the central node into three parts: unique information from each node, redundant information, and synergistic information due to the combined information from nodes. 
We furthermore show that typical MPNNs cannot capture redundant and synergistic information. 

To tackle these limitations we propose the \textit{Graph Ordering Attention (GOAT)} layer, a novel architecture that can capture all sources of information. 
We employ self-attention to construct a permutation-invariant ordering of the nodes in each neighborhood before we pass these ordered sequences to a Recurrent Neural Network (RNN) aggregator. Using a permutation-sensitive aggregator, such as the Long Short-Term Memory (LSTM) model, allows us to obtain larger representational power \citep{murphy2019janossy} and  to capture the redundant and synergistic information. 
We further argue that the ordering of 
neighbors plays a significant role in the final representation \citep{DBLP:journals/corr/VinyalsBK15} and demonstrate the effectiveness of GOAT versus other non-trainable and/or permutation-sensitive aggregators with a random ordering \citep{hamilton2018inductive}.

Our main contributions are summarized as follows:
\begin{enumerate}
    \item We present a novel view of learning on graphs based on information theory and specifically on the Partial Information Decomposition. 
    We further demonstrate that typical GNNs can not effectively capture  redundant and synergistic information between nodes. 
    \item We propose the \textit{Graph Ordering Attention (GOAT)} layer, a novel GNN component that can capture synergistic information between nodes using a recurrent neural network (LSTM) as an aggregator.
   We highlight that the ordering of the neighbors is crucial for the performance and employ a self-attention mechanism to learn it. 
   \item  We evaluate GOAT in node classification and regression tasks on several real-world and synthetic datasets and outperform an array of state-of-the-art GNNs.
\end{enumerate}

\section{Preliminaries and Related Work}
We begin by defining our notation and problem context.  

\textbf{Problem Formulation and Basic Notation.}
Let a graph be denoted by $G = (V,E),$ where $V = \{v_1,\ldots,v_N\}$ is the node set and  $E$ is the edge set. Let $\mA \in \mathbb{R}^{N \times N}$ denote the adjacency matrix, $\mX=[x_1,\ldots,x_N]^T \in \mathbb{R}^{N \times d_I}$ be the node features and $\bm{Y} = [y_1,\ldots,y_N]^T \in \mathbb{N}^{N}$ the label vector. 
We denote the neighborhood of a vertex $u$ by $\neighborhood{u}$ such that
$\neighborhood{u}$ = $\{v : (v, u) \in E\}$ and the neighborhood features by the multiset $\bm{X}_{\neighborhood{u}} = \{x_v : v \in \neighborhood{u}\}$.
We also define the neighborhood of $u$ including $u$ as $\closedneighborhood{u}= \neighborhood{u}\cup \{ u\}$ 
and the corresponding features as $\bm{X}_{\closedneighborhood{u}}$.
The goal of semi-supervised node classification and regression is to predict the labels of a test set given a training set of nodes.

\textbf{Graph Neural Networks.} GNNs exploit the graph structure $\bm{A}$ and the node features $\bm{X}$ in order to learn a hidden representation $h_u$ of each node $u$ such that the label $y_u$ can be predicted accurately from $h_u$ \citep{1555942,4700287}. 
Most 
approaches use a neighborhood message-passing scheme, in which every node updates its representation by aggregating the representations of its neighbors and combining them with its previous representation,
\begin{align*}
    m_u^{(\layersuperscript)} &= \text{Aggregate}^{(\layersuperscript)} \left( \left\lbrace h_v^{(\layersuperscript-1)} : v \in \neighborhood{u} \right\rbrace \right),\numberthis \label{eqn:aggregation}\\
    h_u^{(\layersuperscript)} &= \text{Combine}^{(\layersuperscript)} \left( h_u^{(\layersuperscript-1)}, m_u^{(\layersuperscript)} \right),
\end{align*}
where $h_u^{(\layersuperscript)}$ denotes the hidden representation of node $u$ at the $\layersuperscript^{\mathrm{th}}$ layer of the GNN architecture. Note that we often omit the superscript $(\layersuperscript)$ to simplify the notation. 

Typically GNNs employ a permutation-invariant ``Aggregate'' function to yield a permutation-equivariant GNN layer \citep{bronstein2021geometric}. Permutation invariance and equivariance will be defined formally now. 

\begin{definition}
\label{defn:permutations}
Let $S_M$ denote the group of all permutations of a set containing $M$ elements. A function $f(\cdot)$ is \textit{permutation-equivariant} if for all $\pi \in S_M$ we have $\pi f(\{x_1, x_2, .., x_M\}) = f(\{x_{\pi(1)}, x_{\pi(2)}, \ldots , x_{\pi(M)}\}).$ A function $f(\cdot)$ is \textit{permutation-invariant} if for all $\pi \in S_M$ we have $f(\{x_1, x_2, .., x_M\}) = f(\{x_{\pi(1)}, x_{\pi(2)}, \ldots , x_{\pi(M)}\}).$
\end{definition}

\subsection{Common Aggregators and Their Limitations}\label{sec:aggregator_limitations}

\comment{
\begin{table*}
    \begin{tabular}{|c|c|c|c|c|c}
    \hline
    \textbf{Name} & \textbf{Formula}  & \textbf{Models} & \textbf{Injectivity} & \textbf{Learnable} & \textbf{Relational Reasoning} \\ 
    \hline Max & $max(\{h_u , \forall u \in N_v\})$ & & $\times$  & $\times$  & $\times$  \\
    Min & $min(\{h_u , \forall u \in N_v\})$ & & $\times$  & $\times$  & $\times$  \\
    Mean & $mean(\{h_u , \forall u \in N_v\})$ & GCN \citep{kipf2017semisupervised} & $\times$  & $\times$  & $\times$  \\ 
    Sum & $(1+\epsilon)h_v + \sum_{u \in N_v^*} h_u$ & GIN \citep{xu2019powerful} & \checkmark & $\times$ & $\times$  \\
    Attention & $\sum_{u \in N_v} a_{vu}h_u $ & GAT \citep{velikovi2017graph} & $\times$ & & 2-tuples\\
    \hline
    GOAT & & & TO BE PROVEN & \checkmark & N-tuples
    \end{tabular}
    \caption{Common aggregators used in Graph Neural Networks. We examine if they satisfy 3 basic properties: \textbf{Injectivity},  \textbf{Relational Reasoning.}} 
\end{table*}
}

We now describe some of the most well-known aggregators and  discuss their limitations. 
Our analysis is based on two important properties that an aggregator should have: \begin{enumerate}
\item \textbf{Relational Reasoning}: 
The label of a node may depend not only on the unique information of each neighbor, but also on the joint appearance and interaction of multiple nodes \citep{wagstaff2021universal}. 
With the term ``relational reasoning'' we describe the property of capturing these interactions, i.e., synergistic information, when aggregating neighborhood messages. 
\item \textbf{Injectivity}: As shown in \citet{xu2019powerful}, a powerful GNN should map two different neighborhoods, i.e., multisets of feature vectors, to different  representations. 
    Hence,  the aggregator should be injective.  
\end{enumerate} 

The \textit{mean} and \textit{max} functions are commonly used to aggregate neighborhood information~\citep{kipf2017semisupervised}. However, they are neither injective \citep{xu2019powerful} nor able to perform relational reasoning as they process each node independently. The \textit{summation operator followed by a multilayer perceptron} was recently proposed \citep{xu2019powerful}. This aggregator is injective but cannot perform relational reasoning and usually requires a large latent dimension \citep{wagstaff2019limitations,wagstaff2021universal}.
In the Graph Attention Networks (GAT)~\citep{velikovi2017graph}, the representation of each node is computed  by applying a \textit{weighted summation} of the representations of its neighbors. 
However, the attention function is not injective since it fails to capture the cardinality of the neighborhood.
Recently, an improved version of the GAT was published \citep{brody2022how} and also, a new type of attention was proposed \citep{Zhang_2020}, that preserves the cardinality of the neighborhood and therefore is injective. Nevertheless, none of these models can capture interactions between neighbor nodes as each attention score is computed based only on the representations of the central node and one neighbor node. In Section \ref{subsec:inf_agg} we provide further details on why typical aggregators fail, from an information theoretic perspective.

\subsection{Permutation-Sensitive Aggregators}

Several authors have proposed the use of permutation-sensitive aggregators to tackle the limitations of permutation-invariant aggregators. In particular, \citet{Niepert2016} propose to order nodes in a neighborhood according to some labeling, e.g., the betweeness centrality or PageRank score, to assemble receptive fields, possibly extending beyond the 1-hop neighborhood of a central node, which are then fed to a Convolutional Neural Network (CNN) architecture. While this approach demonstrates good performance, it relies on the fixed chosen ordering criterion to be of relevance in the context of a given dataset and chosen learning task. \citet{Gao2018} propose to only work with the $k$ largest hidden state values for each hidden state dimension in each neighborhood. While not explicitly ordering the neighboring nodes, this operation summarises any given neighborhood in a fixed size feature matrix and enables the use of permutation-sensitive aggreators, CNNs in their case. Of course the choice of $k$ involves a loss of information in almost all cases, i.e., when $k$ is smaller than the maximal degree in the graph. In the Janossy Pooling \citep{murphy2019janossy} approach, a permutation-invariant aggregator is obtained by applying a permutation-sensitive function to all $n!$ permutations. Since the computational cost of this approach is very high, they also propose an approximation, sampling only a limited number of permutations.
Similarly, in the GraphSage \citep{hamilton2018inductive} model, a random permutation of 
each neighborhood is considered and then passed to an LSTM. However, it has been observed that even in the graph domain, where typically no natural ordering of nodes is known,  there exist some orderings that lead to better model performance \citep{DBLP:journals/corr/VinyalsBK15}. Whether these high performance orderings are discovered during the training process is left to chance in the GraphSage and Janossy Pooling models. 
In contrast, our method learns a meaningful ordering of neighbors with low complexity by leveraging the attention scores.


\section{An Information Theory Perspective}
\label{sec:gmi}
In this section, we show how neighborhood dependencies can be encoded in the Partial Information Decomposition framework. This decomposition will motivate us to build a more expressive GNN layer, that is able to capture various interactions among neighborhood nodes.

\subsection{Partial Information Decomposition} 

The domain of information theory provides a well-established framework for measuring neighborhood influence. 
A few graph representation learning results capitalize on information-theoretic tools, either assuming a probability distribution over the feature vectors~\citep{dgi,mi_survey} or over the structural characteristics~\citep{graphentropy,vnestruct}.

The majority of GNNs (including the attention-based models) use an aggregation that does not capture interactions among neighbors. 
Mutual information is a measure that can give us an insight in the omitted informative interactions. 



\begin{definition}
For a given node $u\in V,$ let  $\mH_{\closedneighborhood{u}} = [h_{v_1},\ldots,h_{v_{|\closedneighborhood{u}|}}]\in \R^{|\closedneighborhood{u}|\times d}$ denote the hidden representations of the nodes in $\closedneighborhood{u}$. Then, if we assume that $\mH_{\closedneighborhood{u}}$ and $ h_u$ follow  distributions $p(\mH_{\closedneighborhood{u}})$ and $ p(h_u),$ respectively, the \textit{mutual information} between $h_u$ and $\mH_{\closedneighborhood{u}}$ is defined as

\begin{multline}
        I(h_u;\mH_{\closedneighborhood{u}}) = \iint p\left(h_u, \mH_{\closedneighborhood{u}}\right) \\\log\left(\frac{p\left(h_u,\mH_{\closedneighborhood{u}}\right)}{p(h_u) p\left(\mH_{\closedneighborhood{u}}\right)}\right)dh_ud\mH_{\closedneighborhood{u}}.\label{eq:mi}
\end{multline}
\end{definition}

Following~\citet{pid}, \eqnref{eq:mi} can be decomposed into three components as follows:
\begin{equation}
    I(h_u;\mH_{\closedneighborhood{u}}) = \sum_{v \in \closedneighborhood{u}} U_v + R + S, 
\end{equation}
\begin{itemize}
    \item The \textit{unique information} $U_v$ for all 
    $v\in \closedneighborhood{u}$ corresponds to the information a neighbor carries independently and no other neighbor has, 
    \item The \textit{redundant information} $R$ is the information that can be found overlapping in two or more neighbors and 
    \item The \textit{synergistic information} $S$ expresses the information that can be captured only if we take the interactions among neighbors into account. 
\end{itemize}

\begin{figure}[t!]
    \centering
    \includegraphics[width=.3\textwidth]{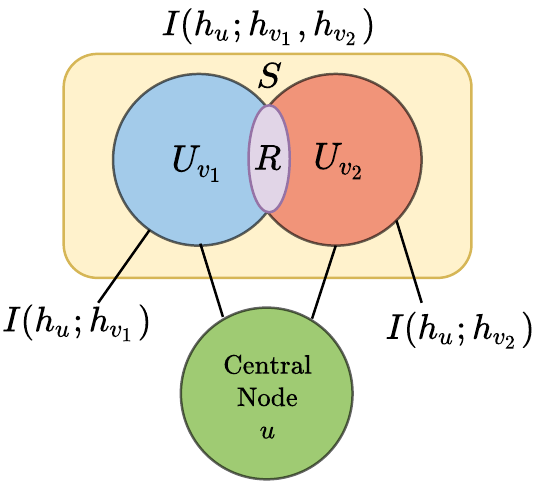}
    \caption{An illustration of the Partial Information Decomposition for the case of one central node $u$ and two neighbors $v_1, v_2$. 
    Each of the mutual information terms $I(h_u;h_{v_1})$ and $I(h_u;h_{v_2})$ consists of the unique information provided by $v_1$ ($U_{v_1}$, blue patch) and  $v_2$ ($U_{v_2}$, red patch), respectively, as well as  the shared information of $v_1$ and $v_2$ ($R$, purple patch).
  The joint mutual information $I(h_u;h_{v_1},h_{v_2})$ (yellow box encompassing the inner two circles) consists of four elements: the unique information in the neighbors $v_1$ and $v_2,$ their redundant information and additionally the synergistic information, $I(h_u;h_{v_1},h_{v_2}) = U_{v_1} + U_{v_2} + R + S.$
    }
    \label{fig:attention_synergy}
\end{figure}

In Figure \ref{fig:attention_synergy}, we provide an illustration of the PID framework. 
To exemplify this concept we discuss it in the context of the much-used Cora dataset, for which node feature vectors contain binary indication of the presence or absence of certain key words in the abstracts of scientific publications \citep{Sen_Namata_Bilgic_Getoor_Galligher_Eliassi-Rad_2008}. For this dataset unique information takes the form of key words, which are present in only one abstract in a given neighborhood, redundant information refers to keywords, which are repeatedly present without their total number of appearances being of consequence to our learning task, and synergistic information refers to insight that can be gained by observing a certain combination of key words.

\subsection{Information Captured by Aggregators}
\label{subsec:inf_agg}
To better understand the information captured by standard GNNs, we first analyze the contribution of each neighboring node to the aggregated representation of a central node.

We assume the structure of an MPNN model, in which each node updates its hidden representation by aggregating information of its neighbors. Further, we denote the message that a given central node $u$ receives from a neighboring node $v$ by $c_{uv} \in \mathbb{R}^d.$ Then, $c_{uv}$ can be interpreted as the \textit{contribution} of node $v$ to the hidden state of $u$ and the aggregated messages in \eqnref{eqn:aggregation} can be expressed as 
$ m_u = \sum_{v \in \neighborhood{u}} c_{uv}.$


\begin{figure*}[t]
    \centering
    \includegraphics[width=0.9\textwidth]{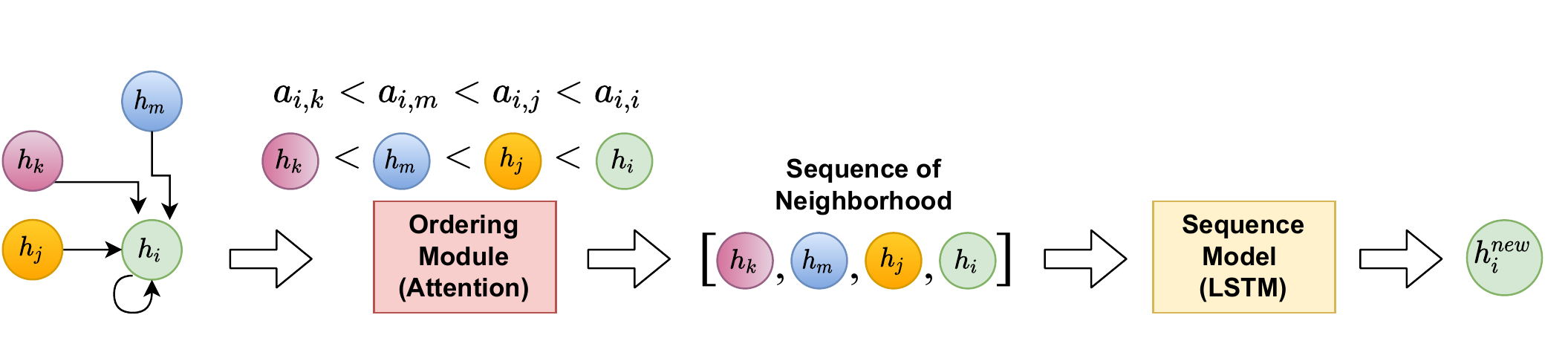}
    \caption{An illustration of the aggregation and update of the representation of node $v_i$ using a GOAT layer. A  self-attention mechanism is used in order to obtain a ranking between the nodes of the neighborhood and then the ordered neighborhood is given as input into a sequence model (LSTM) to produce the updated representation of node $v_i$. }
    \label{fig:example_architecture}
\end{figure*}

For the Graph Isomorphism Network (GIN) \citep{xu2019powerful} and Graph Convolutional Network (GCN) \citep{kipf2017semisupervised} we observe that $c_{uv} = f(\gso_{uv},h_v) = \gso_{uv}h_v$, where $\gso \in \mathbb{R}^{N\times N}$ is a graph shift operator, such as $\mA$ or $(\mD+\mI)^{-1/2}(\mA+\mI)(\mD+\mI)^{-1/2}$, and $\mD$ denotes the graph's degree matrix. The contribution of each neighbor $v$ is only determined by its hidden state $h_v$ and the value $\gso_{uv}$ of the graph shift operator.
For the GAT we observe that $c_{uv} = f(h_u,h_v) = a_{uv}h_v$, where $a_{uv}$ is the attention score that is computed from $h_u$ and $h_v$. The contribution of each neighbor is also affected by the hidden state of the central node, but is not affected by the other neighbors. 

We argue that processing each neighbor individually limits current aggregators, as any interactions among neighbors are ignored by design. Therefore, they can not capture synergistic information between nodes, i.e., the amount of information that is captured equals $\sum_{v \in \neighborhood{u}} I(h_u;h_v).$ Consider the example of a neighborhood with two neighbors $v_1,v_2$. The information captured by a standard GNN is expressed in terms of the PID as follows, $I(h_u;h_{v_1})+I(h_u;h_{v_2}) = U_{v_1} + U_{v_2} + 2R,$ which is different from the joint mutual information $I(h_u;h_{v_1},h_{v_2}) = U_{v_1} + U_{v_2} + R + S.$ Thus, the captured information from a standard GNN is less than the information present in the neighborhood due to the absence of synergistic information.

To address this problem, we introduce a dependence of the contribution $c_{uv}$ of the neighbor node $v$ 
on all neighbors of $u$. Therefore, $c_{uv}$ is now a function not only of $h_u$ and $h_v,$ but also of $h_j$ for $ j \in \neighborhood{u}$, i.e., $c_{uv} = f(\gso,\mH_{\closedneighborhood{u}}).$ To achieve this, we learn a meaningful ordering of the neighbor nodes using an attention mechanism, and then use an RNN to aggregate the representations of the neighbors. 

\section{Graph Ordering Attention Layer}
\label{sec:GOAT_architecture}


We now present the architecture of our \textit{Graph Ordering Attention (GOAT)} layer and highlight its theoretical advantages over other message-passing models. A deep GNN can be constructed by stacking several GOAT layers or combining GOAT layers with other GNN layers. A GOAT layer (illustrated in Figure \ref{fig:example_architecture}) consists of two parts:


1) The \textbf{Ordering Part} (red box in Figure \ref{fig:example_architecture}) 
transforms the unordered multiset of neighbor hidden state vectors, each of dimension $d,$ $\{h_1,\ldots,h_{\cnsize}\},$ with $\cnsize = \left|\closedneighborhood{u}\right|,$ into an ordered sequence, using an attention mechanism,
\begin{equation*}
    [h_{\pi(1)},\ldots,h_{\pi(\cnsize)}] = \text{OrderingPart}(\{h_1,\ldots,h_{\cnsize}\}),
\end{equation*}
where the ordering is given by the permutation function $\pi(\cdot).$

Specifically, similar to the GAT \citep{velikovi2017graph} model, for each node $v_i \in V$, we first apply a shared linear transformation parameterized by a \emph{weight matrix} ${\bf W_1} \in \mathbb{R}^{d \times d}$ and then perform a shared self-attention mechanism parameterized by $\vec{w}_2 \in \mathbb{R}^{2d}$ to compute the \emph{attention scores} 
\begin{equation} 
\label{eq:attenton_coeff}
	a_{ij} = \text{LeakyReLU}\left(\vec{w}_2^T[{\bf W_1}{h}_i\|{\bf W_1}{h}_j]\right),
\end{equation} 
for all 
$j$ such that $v_j \in \closedneighborhood{v_i}$. 
Then, we sort the coefficients in decreasing order of magnitude
\begin{equation}
\label{eq:sorting}
  a_{i\pi(1)},\ldots,a_{i\pi(\cnsize)} =  sort\left(a_{i1},\ldots,a_{i\cnsize}\right),
\end{equation}
obtaining a specific permutation $\pi$ of the nodes in the neighborhood. 
When all attention scores are different from each other, we observe that the sorting function in \eqnref{eq:sorting} is deterministic and permutation invariant. 
In cases where two or more nodes have equal attention scores, we resort to an additional sorting criterion, described in Appendix \ref{app:sorting_criterion},  to ensure that our sorting function is deterministic and permutation invariant.


Once we obtain the permutation $\pi$, we construct the sorted sequence of neighbourhood hidden states
\begin{multline} 
\label{eqn:h_sorted}
    h_{sorted(i)} = \left[\frac{e^{a_{i\pi(1)}}}{\sum_{j=1}^Qe^{a_{i\pi(j)}}}{\bf W_1}h_{\pi(1)}, \dots,\right.\\ 
    \left.\frac{e^{a_{i\pi(Q)}}}{\sum_{j=1}^Qe^{a_{i\pi(j)}}}{\bf W_1}h_{\pi(\cnsize)}\right].
\end{multline}

Note that we use the attention scores to both order the hidden states and, after normalisation via the softmax function, as coeffiecients for the hidden states.  Only due to the occurrence of the attention coefficients in \eqnref{eqn:h_sorted} are we able to obtain gradients in the backpropagation algorithm for ${\bf W_1}$ and $\vec{ w}_2$ (the sorting function in \eqnref{eq:sorting} is not differentiable). 
Note further that any self-attention mechanism, such as the GATv2 by \citet{brody2022how}, can be used instead of GAT, to obtain the attention scores in a GOAT layer. 

2) The \textbf{Sequence Modeling Part} (yellow box in Figure~\ref{fig:example_architecture}) takes the ordered sequences of nodes produced by the Ordering Part  as input and processes them using an RNN, that is shared across all neighborhoods, to generate the new hidden states. In the PID context, the Bidirectional LSTM  \citep{10.1162/neco.1997.9.8.1735} appears to be the best suited RNN available. 
Its forget gate allows us to discard redundant information; the input gate is sufficiently expressive to isolate unique information; while its memory states allow for the identification of synergistic information. 
\begin{equation}
    h^{new}_i = \text{LSTM}\left(h_{sorted(i)}\right) \in \mathbb{R}^{d_O}.
\end{equation}
Since we utilize a Bidirectional LSTM the contribution of each node, discussed in Section \ref{subsec:inf_agg}, depends on all other hidden states in the neighborhood. 
Specifically, each direction in the Bidirectional LSTM ensures that both the nodes preceding and succeeding a particular node $j$ are taken into account when calculating the contribution $c_{ij}$  of node $j.$ 


Note that the choice of the LSTM is made without loss of generality and any RNN could be chosen in the Sequence Modeling Part. Indeed, in Section \ref{sec:experiments} we will observe results for a variety of RNNs chosen as part of our GOAT layer. To work with a faster, more scalable implementation we pad all neighborhood sequences with zero vectors to be equal in length to the sequence of hidden states arising in the largest neighborhood in the graph. This allows us to train the LSTM on larger batches of neighborhoods in parallel. The alternative implementation, where neighborhood sequences of different length are fed to the LSTM individually is an equally valid, while slower, implementation.

\textbf{Multi-Head Attention Ordering.} We can also employ multi-head attention to provide additional representational power to our model. We see several advantages in the consideration of multiple heads in our architecture. If only one sensible ordering of the nodes in a neighborhood exists, then multiple heads can help us estimate this ordering more robustly. If on the other hand there exist several sensible orderings of the nodes in a neighborhood, then a multi-head architecture allows us to take all of these into account in our model.
Let $K$ be the number of the attention heads.  
Equation \eqnref{eq:attenton_coeff} for the $k$-th attention head is transformed as
\begin{equation*}
    a_{ij}^k = a^k({\bf W_1^k}{h}_i, {\bf W_1^k}{h}_j).
\end{equation*}
Then we sort the $K$ sets of attention scores obtaining multiple orderings of the neighborhood, $h_{sorted(i)}^k$ for $k \in \{1, \ldots, K\}.$
To generate the final representation of the nodes we concatenate the features from the $K$ independent Bidirectional LSTM models, i.e., 
\begin{equation*}
    h^{new}_i = \concat_{k=1}^K \text{LSTM}^k\left(h_{sorted(i)}^k\right).
\end{equation*}

\textbf{Complexity.} The time complexity, derived in Appendix \ref{app:complexity}, of a single-head GOAT layer is $\mathcal{O}(|V| d_O  d+ |E| d_O + |V| d_{\max} \log(d_{\max}) + |V|  d_{\max} 4  d(d_O + d+3)),$ where $d_{\max}$ denotes the maximal degree in the graph. 
For $d_{\max}\ll d,$ the only additional complexity introduced by our model manifests in the multiplicative $d_{\max}$ term in the last summand of the complexity expression. Limiting the maximal degree by applying a sampling strategy, limits the additional complexity our model introduces. Hence, the time complexity of our GOAT model can be comparable to standard MPNNs models. 
Note that the space complexity of a GOAT layer only exceeds the space complexity of a GAT layer by the space complexity of the LSTM model. 


\textbf{Permutation-Equivariance and Injectivity of GOAT.}
Recall from Section \ref{sec:aggregator_limitations} that the permutation-equivariance and injectivity are desirable properties for a GNN layer to have. We will now prove that our GOAT layer satisfies both of these criteria.

\begin{proposition}[Permutation-Equivariance of GOAT]
\label{prop:permutation_equivariance}
Our GOAT layer performs a permutation-equivariant transformation, with respect to node label permutations, of the hidden states corresponding to the nodes in a graph.
\end{proposition}
The proof of Proposition \ref{prop:permutation_equivariance} is in Appendix \ref{app:permutation}. Therefore, our GOAT layer is able to benefit from the expressivity of a permutation-sensitive aggregator, while acting on the node's hidden representations in a permutation-equivariant way.

Our result on the injectivity of a GOAT layer relies on the concept of function approximation in probability. We reproduce the definition of this concept from \citet{Hammer2000}. 

\begin{definition} \label{defn:approx_in_prob}
Let $\mathcal{X}$ denote the space of finite lists with elements in $\mathbb{R}^q$ for $q \in \mathbb{N}$ and $P$ be a probability measure on $\mathcal{X}.$  For measurable functions $f_1,f_2:\mathcal{X}\rightarrow \mathbb{R}^t$ we say that $f_1$ \textit{approximates} $f_2$ with accuracy $\epsilon>0$ and confidence $\delta>0$ \textit{in probability} if $P(x\in \mathcal{X}\mid |f_1(x) - f_2(x)|>\epsilon)<\delta.$
\end{definition}

\begin{theorem}[Injectivity of GOAT]
\label{thm:injectivity}
Assume that for all nodes $u \in V$ the multisets of hidden states corresponding to its neighbors is finite and has elements in $\mathbb{R}^q$ for $q \in \mathbb{N}.$ Then, there exists a GOAT layer approximating a measurable function arbitrarily well in probability for which any two distinct multisets are mapped to distinct node representations. 
\end{theorem}
Hence, we have shown that our GOAT layer is sufficiently expressive to approximate any measurable injective function in probability.  The proof of Theorem \ref{thm:injectivity} is in Appendix \ref{app:injectivity}. 

\begin{table*}[t!]
\centering
\caption{Classification accuracy ($\pm$ standard deviation) on the ``Top-2 pooling'' synthetic dataset and MSE ($\pm$ standard deviation) results on the synthetic datasets ``Betweenness Centrality'' and ``Effective Size'' for two different types of random graphs.}
\label{exp:structural_properties}
\resizebox{1.7\columnwidth}{!}{
\begin{tabular}{lrrrrrr}
\toprule
\multirow{2}{*}{\textbf{Method}} &
\multicolumn{1}{c}{\textbf{Top-2 pooling}} &  \multicolumn{2}{c}{\textbf{Betweenness Centrality \footnotesize{(MSE)}}} &  \multicolumn{2}{c}{\textbf{Effective Size \footnotesize{(MSE)}} } \\ & \textbf{\footnotesize{(Accuracy)}} & N=100, p=0.09 & N=1000, p=0.01 & N=100, p=0.09 & N=1000, p=0.01 \\ 
\midrule
GCN & 57.35 $\pm 4.13$ & 0.0063 $\pm 0.0036$ & 0.0020 $ \pm 0.0008$ & 0.0135 $\pm 0.0067 $ & 0.00380 $\pm 0.00120 $ \\
GraphSAGE (mean)  & 61.45 $\pm 5.79$ & 0.0401 $\pm 0.0158$ & 0.0221 $\pm 0.0069$ & 0.0374  $\pm 0.0085$ &  0.02430  $\pm 0.00560 $ 
 \\
 GraphSAGE (lstm) & 65.05 $\pm 8.71$ & 0.0094 $\pm 0.0073$ & 0.0153 $\pm 0.0105$ & 0.0022 $\pm 0.0017 $ &  0.00080 $\pm 0.00020$  \\
GIN & 56.40 $\pm 5.26$ & 0.0083 $\pm 0.0052 $ & 0.0042 $\pm 0.0015$ & 0.0024 $ \pm 0.0016$ & 0.00070 $\pm 0.00030$ \\
GAT & 53.34  $\pm 2.43$ & 0.0409 $\pm 0.0158$ & 0.0220 $ \pm 0.0068$ & 0.0382  $\pm 0.0079$ & 0.02480  $\pm 0.00560$  \\
PNA & 61.50 $\pm 10.9$ & 0.0115 $\pm 0.0089$ & 0.0020 $\pm 0.0008$ & 0.0121 $\pm0.0119$ & 0.00137 $\pm 0.00035$\\
 \hline
\textbf{GOAT(lstm)}  & \textbf{69.21} $\pm 5.10$ & \textbf{0.0038 $\pm 0.0019$} & \textbf{0.0006} $\pm 0.0002$ & \textbf{0.0016} $\pm 0.0008$ & \textbf{0.00020} $\pm 0.00008$ \\ \bottomrule
\end{tabular}}
\end{table*}

\section{Experimental Evaluation}
We perform an extensive evaluation of our GOAT model and compare against a wide variety of state-of-the-art GNNs, on three synthetic datasets (in Sections \ref{sec:exp_syn1} and \ref{sec:exp_syn2}) as well as on nine node-classification benchmarks (in Section \ref{sec:experiments}). Our code is available on \href{https://github.com/MichailChatzianastasis/GOAT}{\color{blue}{github}}.

\textbf{Baselines.} We compare GOAT against the following state-of-the-art GNNs for node classification: 
1)~GCN \citep{kipf2017semisupervised} the classical graph convolution neural network, 
2)~GraphSAGE(mean) \citep{hamilton2018inductive} that aggregates by taking
the elementwise mean value, 
3)~GraphSAGE(lstm) \citep{hamilton2018inductive} that aggregates by feeding the neighborhood hidden states in a random order to an LSTM,
4)~GIN \citep{xu2019powerful} the injective summation aggregator,
5)~GAT \citep{velikovi2017graph} that aggregates with a learnable weighted summation operation,  
6)~PNA \citep{corso2020pna} that combines multiple aggregators with degree-scalers. 
We also compare with 7)~a standard MLP that only uses node features and does not incorporate the graph structure. To better understand the impact of the choice of RNN in the GOAT architecture, we provide results from three different GOAT architectures, in which the standard RNN, GRU \citep{cho-etal-2014-properties} and LSTM are used.

\textbf{Setup.} For a fair comparison we use the same training process for all models adopted by \citet{velikovi2017graph}. We use the Adam optimizer \citep{kingma2017adam} 
with an initial learning rate of 0.005 and early stopping for all models and datasets. 
We perform a hyperparameter search for all models on a validation
set. The hyperparameters include the size of hidden dimensions, dropout, and number of attention heads for GAT and GOAT. We fix
the number of layers to 2. 
In our experiments we combine our GOAT layer with a GAT or GCN layer to form a 2-layer architecture. 
More information about the datasets, training procedure, and hyperparameters of the models are in Appendix~\ref{appendix:datasets}. 

\subsection{Top-2-Pooling} \label{sec:exp_syn1}

In this task, we sample Erdős–Rényi random graphs with 1000 nodes and a probability of edge creation of 0.01. We draw 1-dimensional node features from a Gaussian Mixture model with three equally weighted components with means $1, 1$ and $2$ and standard deviations $1, 4$ and $1.$ 
Then, we label each node with a function $\phi(\cdot, \cdot)$ of the two 1- or 2-hop neighbors that have the two different largest features, i.e., to each node $u \in V$ we assign a label $y_u = \phi(x_a,x_b),$ where $x_a$ and $x_b$ are the largest, distinct node features of all nodes in the 2-hop neighborhood of $u$ with nodes features at a distance of 2 being down-weighted by a factor of 0.8. We set  $\phi$ to be  $\phi(x_a,x_b) = \sqrt{exp(x_a)+exp(x_b)}$. 
Finally, to transform this task to node classification we bin the $y$ values into two equally large classes. We use 60/20/20 percent of nodes for training, validation and testing.

We report the average classification accuracy and the standard deviation across 10 different random graphs in Table \ref{exp:structural_properties}.
\textit{Our GOAT model outperforms the other GNNs with a large margin.} Specifically, our model leads to an 18.36\% increase in performance over GAT and 6.65\% increase over GraphSage(lstm).  
In the context of this simulation study, we explain this performance gap with the following hypothesis. 
To find the largest element of a set one must consider 2-tuple relationships therefore synergistic information is crucial for this task.
An LSTM can easily perform the necessary comparisons with a 2-dimensional hidden space.
As nodes are processed they can either be discarded via the forget gate, if they are smaller than the current hidden state, or the hidden state is updated to contain the new feature node.
In contrast, typical GNNs need exponentially large hidden dimensions in order to capture the necessary information as they cannot efficiently discard redundant information. We observe that GraphSage(lstm) is the second-best performing model due to its LSTM aggregator. However, it does not learn a meaningful ordering of the nodes that simplifies this task.

\begin{table*}[t]
\centering
    \caption{Node classification accuracy using different train/validation/test splits. We highlight the best performing model and underline the second best. Since there exists a single standardised split for Cora and CiteSeer no standard deviations are given.}
    \label{results:node_benchmarks}
    \resizebox{1.8\columnwidth}{!}{
    \begin{tabular}{lcccccccc}
    \toprule
    \textbf{Method} & \textbf{Cora}  & \textbf{CiteSeer} & \textbf{Disease} & \textbf{LastFM Asia} & \textbf{Computers} &
    \textbf{Photo} &  
    \textbf{CS} &
    \textbf{Physics}
    \\
    \midrule
    MLP & 43.8 & 52.9 & 79.10 $ \pm 0.97 $ & 72.27 $\pm 1.00$ & 79.53 $\pm 0.66$ & 87.89 $\pm 1.04$ & 93.76 $\pm 0.26$ & 95.85 $\pm 0.20$ \\
    GCN & 81.4 & 67.5 & 88.98 $ \pm 2.21 $ & \underline{83.58} $\pm 0.93$ & 90.72 $\pm 0.50$ &  93.99 $\pm 0.42$ & 92.96 $\pm 0.32$ & 96.27 $\pm 0.22$\\
    GraphSAGE (mean) & 77.2 & 65.3 & 88.79 $\pm 1.95 $ & 83.07 $\pm 1.19$ & \underline{91.47} $\pm 0.37$ & 94.32 $\pm 0.46$ & \underline{94.11} $\pm 0.30 $ & 96.31 $\pm 0.22$ \\
    GraphSAGE (lstm)  & 74.1 & 59.9 & 90.50 $\pm 2.15$ & \textbf{86.85} $\pm 1.07$ &  91.26 $\pm 0.51$ & 94.32 $\pm 0.64$ & 93.46 $\pm 0.29$ & 96.40 $\pm 0.16$ \\
    GIN & 75.5 & 62.1 & 90.20 $\pm 2.23$ & 82.94 $\pm 1.25$ & 84.68 $\pm 2.33$ & 90.07 $\pm 1.19$ & 92.38 $\pm 0.38$ & 96.38 $\pm 0.16$ \\
    GAT & 83.0 & 69.3 & 89.13 $ \pm 2.22$ & 77.57 $\pm 1.82$  & 85.41 $\pm 2.95 $ & 90.30 $\pm 1.76$ & 92.78 $\pm 0.27$ & 96.17 $\pm 0.18$ \\
    PNA & 76.4 & 58.9 & 86.84 $\pm1.89$ & 83.24 $\pm 1.10$ & 90.80 $\pm 0.51$ & \underline{94.35} $\pm0.68$ & 91.83 $\pm 0.33$ & 96.25 $\pm 0.21$ \\
    \hline
    \textbf{GOAT(lstm)} & \textbf{84.9} & \underline{69.5} & \textbf{92.11} $\pm 1.88$ & 83.29 $\pm 0.91$ & 91.34 $\pm 0.50$ & \textbf{94.38} $\pm 0.66$  & \textbf{94.21} $\pm 0.42$ & \textbf{96.69} $\pm 0.31$ \\
    \textbf{GOAT(gru)} & 83.5 &  \textbf{70.0} & \underline{91.97} $\pm 1.90$ & 83.35 $\pm 0.91$ & \textbf{91.54} $\pm 0.48$ & 94.22 $\pm 0.58$ & 93.62 $\pm 0.22$ & 96.32 $\pm 0.24$ \\
    \textbf{GOAT(rnn)} & \underline{84.2} & 67.9 & 91.67 $\pm 1.69$ & 83.21 $\pm 0.98$ & 89.10 $\pm 0.51$ & 92.45 $\pm 0.60$ & 93.48 $\pm 0.19$ & \underline{96.44} $\pm 0.20$ \\ 
    \bottomrule
    \end{tabular}
    }
\end{table*}

\subsection{Prediction of Graph Structural Properties} \label{sec:exp_syn2}

The experiments in this section establish the ability of our GOAT model to predict structural properties of nodes. 
The first task is to predict the \textit{betweenness centrality} of each node and the second task is to predict the \textit{effective size} of each node. Both of these metrics, defined in Appendix \ref{appendix:datasets}, are affected by the interactions between the neighbor nodes, so synergistic information is crucial for these tasks. We set the input features to the identity matrix, i.e., $\mathbf{X}=\mathbf{I}$ and use two parameter settings to sample Erdős–Rényi random graphs, namely $(N,p) \in \{(100,0.09),(1000,0.1)\}$, where $N$ is the number of nodes and $p$ is the probability of edge creation.
We use $60\%$ of nodes for training, $20\%$ for validation and $20\%$ for testing. 
We train the models by minimizing the Mean Squared Error (MSE).

We report the mean and standard deviation accuracy and MSE across 10 graphs of each type in Table \ref{exp:structural_properties}. 
Our model outperforms all models in both tasks and in both graph parameter settings. GOAT can capture the synergistic information between the nodes, which is crucial for predicting the betweenness centrality and effective size. The other aggregators miss the structural information of nodes in neighborhoods. We observe that GraphSAGE(lstm) that uses a random node ordering is not on par with GOAT, indicating that the learned ordering in GOAT is valuable here also. 

\subsection{Node Classification Benchmarks} \label{sec:experiments}
We utilize nine well-known node classification benchmarks to validate our proposed model in real-world scenarios originating from a variety of different applications. Specifically, we use 3 citation network benchmark datasets: Cora, CiteSeer~\citep{Sen_Namata_Bilgic_Getoor_Galligher_Eliassi-Rad_2008}, ogbn-arxiv \citep{ogb}, 1 disease spreading model: Disease~\citep{chami2019hyperbolic}, 
1 social network: LastFM Asia~\citep{feather}, 2 co-purchase graphs: Amazon Computers, Amazon Photo~\citep{shchur2019pitfalls} and 2 co-authorship graphs: Coauthor CS, Coauthor Physics~\citep{shchur2019pitfalls}. For the GOAT results on the ogbn-arxiv dataset we randomly sample 100 neighbors per node to represent the neighborhoods for faster computation. 
We report the classification accuracy results in Tables \ref{results:node_benchmarks} and \ref{tab:ogb}. Our model outperforms the others in eight of nine datasets. This demonstrates the ability of GOAT to capture the interactions of nodes, that are crucial for real-world learning tasks. 

\begin{table}[t]
    \centering
       \caption{Node classification accuracy on the ogbn-arxiv dataset. We used the same setup and the reported results from \citet{https://doi.org/10.48550/arxiv.2204.04879}.} 
       \small
    \begin{tabular}{lc}
        \toprule
        \textbf{Method} & \textbf{ogbn-arxiv} \\
        \midrule
         GCN & 33.3 $\pm 1.2$\\
         GraphSAGE & 54.6 $\pm 0.3 $\\
         GAT & 54.1 $\pm 0.5 $\\
         \hline
         \textbf{GOAT(lstm)} & \textbf{55.1} $\pm 0.4$ \\
         \bottomrule
    \end{tabular}
    \label{tab:ogb}
\end{table}

\subsection{Ablation Studies on the Learned Ordering.} 
\label{sec:fixed_ordering}

In our GOAT architecture we make the implicit assumption that ordering neighborhoods by the magnitude of the trainable attention scores is an ordering that results in a well-performing model. We now perform several ablation studies where we compare the GOAT model to models with fixed neighborhood node orderings (GOAT-fixed). 

\textbf{Setup.} We train our GOAT model on the Cora and Disease datasets, using $8$ attention heads and the LSTM aggregator. We store the ordering of the nodes in each neighborhood for each attention head for each epoch. 
Then, we train various (GOAT-fixed) models that use different fixed orderings extracted from the initial model. 
Specifically, we train $4$ different (GOAT-fixed) models with orderings extracted from the $0,100,200,500$ epochs respectively. 
We run the experiment $3$ times 
and report the results in Table \ref{table:fixed_ordering}.

\textbf{Discussion.} We observe that GOAT-fixed-0, which uses a random ordering, since the ordering is extracted before training, achieves the worst performance. 
This highlights the importance of a meaningful ordering of the nodes, and the ability of our model to learn one.
We also observe that the fixed ordering extracted from epoch $100$ outperforms the GOAT model. 
We believe that this phenomenon is associated with the training dynamics of our model. 
Having a fixed ordering may lead to more stability in the learning of high-performing model parameters not associated with the ordering.
For practitioners, learning an ordering in a first run of our model and then training with an extracted fixed ordering may therefore be most advisable.

\begin{table}[t]
\centering
    \caption{Accuracy of the GOAT model using fixed orderings extracted from different epochs of the baseline model's training.}
    \label{table:fixed_ordering}
        \small
    \begin{tabular}{lcccc}
    \toprule
    \textbf{Method} & \textbf{Cora}  & \textbf{Disease} \\
    \midrule
    GOAT & 83.36 $\pm 0.42$ & 90.64 $\pm 0.40$ & \\
    GOAT-fixed-0 & 81.56 $\pm 0.19$ & 90.48 $\pm 0.29$ \\ 
    GOAT-fixed-100 & \textbf{83.80} $\pm 1.14$ & \textbf{90.90} $\pm 0.26$ \\
    GOAT-fixed-200 & 82.13 $\pm 0.27$ & 90.57 $\pm 0.59$ \\
    GOAT-fixed-500 & 82.27 $\pm 0.34$ & 90.50 $\pm 0.49$ \\
    \bottomrule
    \end{tabular}
\end{table}



\textbf{Additional Ablation Studies} 
In Appendix \ref{app:gat_vs_gatv2} we investigate the potential use of the GATv2 model instead of the GAT model in a GOAT layer and find that the two model variants perform comparably. In Appendix \ref{app:janossy} we observe the GOAT model to significantly outperform the Janossy Pooling approach on the Cora, CiteSeer and Disease datasets. In Appendix \ref{app:att_heads} we find the optimal number of attention heads in a GOAT layer to be related to complexity of the learning task on a given dataset. In particular, we observe one attention head to yield optimal performance on Cora, four attention heads are optimal for CiteSeer, while eight attention heads resulted in the best performing model in the Disease dataset. 

\section{Conclusion}
We have introduced a novel view of learning on graphs by introducing the Partial Information Decomposition to the graph context. 
This has allowed us to identify that current aggregation functions used in GNNs often fail to capture synergistic and redundant information present in neighborhoods. 
To address this issue we propose the Graph Ordering Attention (GOAT) layer, which makes use of a permutation-sensitive aggregator capable of capturing synergistic and redundant information, while maintaining the permutation-equivariance property. The GOAT layer is implemented by first learning an ordering of nodes using a self-attention and by then applying an RNN to the ordered representations. This theoretically grounded architecture yields improved accuracy in the node classification and regression tasks on both synthetic and real-world networks. 


\comment{
\begin{table*}
    \begin{tabular}{|c|c|c|c|}
    \hline
    \textbf{Method} & \textbf{Cornell}  & \textbf{Texas} & \textbf{Wisconsin} \\
    \hline
    MLP & \textbf{76.22} $\pm 5.51$ & \textbf{76.22} $\pm 5.51$ & \underline{75.29} $\pm 5.63$\\
    \hline
    GCN & 52.16 $\pm 8.20$ & 56.49 $\pm 8.83$ & 48.43 $\pm 4.38$ \\
    GraphSAGE (mean) & 68.65 $\pm 7.57$ & 71.62 $\pm 10.69$ & 76.02 $\pm 6.17$ \\
    GraphSAGE (lstm) & 63.51 $\pm 9.53 $ & 67.02 $\pm 5.89 $ & 74.11 $\pm 6.12$ \\
    GIN & 51.62 $\pm 7.68$ & 53.24 $\pm 8.55$ & 50.59 $\pm 7.98$  \\
    GAT & 54.32 $\pm 9.55$ & 53.51 $\pm 6.7$ & 48.43 $\pm 7.18$ \\
    \hline
    \textbf{GOAT} & \underline{71.35} $\pm 7.56$ & \underline{74.59} $\pm 7.85$ & \textbf{76.08} $\pm 6.31$  \\
    \end{tabular}
    \caption{Results in terms of classification accuracies in non-homophilous datasets}
\end{table*}
}

\bibliographystyle{unsrtnat}
\bibliography{goat}


\clearpage
\appendix
\section*{Appendix of Graph Ordering Attention Networks}
\setcounter{section}{0}
\section{Resolving the Ties in Attention Scores}
\label{app:sorting_criterion}
When several neighboring nodes have equal attention scores in a given neighborhood, then a simple ordering by attention scores is not deterministic and permutation-invariant. Therefore, we introduce an additional sorting criterion to resolve the ties between equal attention score nodes. Specifically, in this additional sorting criterion we compare the hidden state elements of nodes successively until we detect unequal elements, which are then used to order the nodes in ascending order by discriminating element. In Algorithm \ref{alg:resolve_ties} we lay out the necessary steps in pseudocode. 

For example, assume nodes $v_j,v_k$ in the neighborhood of central node $v_i$ are assigned equal attention scores, but have different embedding vectors $h_j,h_k \in \mathbb{R}^d$,  $h_j \neq h_k$,  
\begin{align*}
    h_j &= \left[ h_{j1}, h_{j2}, \ldots, h_{jd}\right],\\
    h_k &= \left[ h_{k1}, h_{k2}, \ldots, h_{kd}\right].
  \end{align*} 
Then, there exists at least one index $\ell \in \{1, \ldots,d\}$ such that $h_{j\ell} \neq h_{k\ell}$, since $h_j \neq h_k$. We consider the smallest index $\ell$ for which $h_{j\ell} \neq h_{k\ell}.$ If $h_{j\ell} < h_{k\ell}$ then we put $v_j$ first in our learned ordering, otherwise we put $v_k$ first.
To illustrate this further, if $h_j = [1,~ 2,~ 3]$ and  $h_k = [1,~ 3,~ 3]$ and $v_j,v_k$ have equal attention scores, then $v_j$ will be put ahead of $v_k$ in our ordering, since $h_{j2} = 2 < 3 = h_{k2}$. 
This sorting criterion also resolves ties of more than two nodes. Therefore, our sorting function is deterministic and permutation-invariant, even for the case of equal attention scores between two or more nodes.
 


\begin{algorithm}[h]
   \caption{Resolve Ties}
   \label{alg:resolve_ties}
\begin{algorithmic}[1]
   \STATE {\bfseries Input:} Nodes $v_j, v_k$ with embedding vectors $h_j,h_k \in \mathbb{R}^d$ and equal attention scores $a_{ij},a_{ik}, a_{ij} = a_{ik}$.
   \FOR{$q=1$ {\bfseries to} $d$}
    \IF{$h_{jq} < h_{kq}$} 
        \STATE first $\gets v_j$
        \STATE second $\gets v_k$
        \BREAK
    \ELSIF{$h_{jq} > h_{kq}$}
        \STATE first $\gets v_k$
        \STATE second $\gets v_j$
        \BREAK
    \ENDIF
    \ENDFOR
   \STATE {\bfseries Output:} first, second
\end{algorithmic}
\end{algorithm}

\section{Computational Complexity}
\label{app:complexity}
Our GOAT model requires the computation of the three following steps: 
\begin{enumerate}
\item \textbf{Computation of attention scores}. The computational complexity of the GAT or GATv2 model, used to calculate the attention scores,  is $\mathcal{O}(|V|  d_O d  + |E| d_O)$, where $d$ is input dimensions and $d_O$ is the output dimensions \citep{brody2022how}. 
\item \textbf{Sorting attention scores}. The sorting operation requires $\mathcal{O}(d_u  \log(d_u))$ steps for each node $u,$ the degree of which we denote by $d_u.$ To parallelize the computation across the nodes, we pad all the neighborhoods to the maximum degree in the graph $d_{\max}$. Therefore, the complexity of the second step is $\mathcal{O}(|V| d_{\max})  \log(d_{\max})))$.
\item \textbf{Update hidden states using an RNN}.  The total number of parameters in a standard LSTM network is equal to $W = 4 h (d_O + h)   + 4h$, where $d_O$ is the number of input units, and $h$ is the number of output units. The evaluation of the 5 activation funcations and 3 element-wise products involves a total complexity of $\mathcal{O}(8h)$
Therefore, the computational complexity per time step is $\mathcal{O}(4 h (d_O + h + 3))$.  
So, the complexity of computing the representation of each node is equal to $\mathcal{O}( d_u 4 h (d_O + h + 3))$. 
Since we parallelize the computation across the graphs by using the padded sequence, we end up with complexity $\mathcal{O}(|V|  d_{\max}4 h (d_O + h + 3))$.
\end{enumerate}
Therefore, the final complexity of our model is 
\begin{multline*}
\mathcal{O}(|V|  d_O  d  + |E|  d_O + |V|  d_{\max}  \log(d_{\max}) \\+ |V|    d_{\max} 4 h (d_O + h + 3)).
\end{multline*}

If the maximal degree $d_{\max}$ of the graph is large, we can apply a neighbor sampling strategy like in the GraphSAGE model, instead of working with the whole neighborhood. This allows us to assume that $d_{\max} \ll d,d2.$ In this case our complexity is $\mathcal{O}(|V| d_O d +|E| d_O + |V| W  d_{\max}).$ Limiting the maximal degree, limits the additional complexity our model introduces. 


\section{Proof of Proposition \ref{prop:permutation_equivariance}}\label{app:permutation}

 Typically GNNs construct permutation-equivariant functions on graphs by applying a permutation-invariant local function 
over the neighborhood of each node \citep{bronstein2021geometric}. 
To establish the permutation-equivariance of the GOAT layer it therefore suffices to show that the node-wise operation performed by our GOAT layer is permutation-invariant. To do so we make use of the following proposition which concerns the permutation-invariance of composed functions.  
\begin{proposition}
\label{composition_perm_inv}
For any function $f: X \to Y$ and for any permutation-invariant function $g: Z \to X$,  their composition $f \circ g$ is permutation-invariant.
\end{proposition}

Now since the GOAT layer is formed by the composition of the Sequence Modelling Part and the Ordering Part, by Proposition \ref{composition_perm_inv} it suffices to show that the Ordering Part is permutation-invariant to establish the permutation-invariance of their composition. Recall, that in the Ordering Part of the GOAT layer  we implement an attention mechanism on the hidden states of the central node and each neighboring node. Then, nodes are ordered according to the magnitude of the attention scores. Crucially, these computations are independent of the node labelling, even when equal attention scores arise as we show in Appendix \ref{app:sorting_criterion}, making the Ordering Part of the GOAT layer permutation-invariant. Consequently, we apply a local permutation-invariant function rendering the action of the GOAT layer on the graph permutation-equivariant.

\begin{table*}[t]
\begin{center}
\caption{Summary of the datasets used in our experiments.}\label{datasets}
\begin{tabular}{lcccccccc}
\toprule 
 & {\bf Cora} & {\bf CiteSeer} & {\bf Disease} & {\bf LastFM Asia} & {\bf Computers} & {\bf Photo} & {\bf CS} & {\bf Physics}  \\ \midrule
{\bf \# Nodes} & 2708  & 3327 & 1044 & 7624 & 13752 & 7650 & 18333 & 34493 \\
{\bf \# Edges} & 5429 & 4732 & 1043 & 27806 & 491722 & 238162 & 163788 & 495924 \\
{\bf \# Features/Node} & 1433 & 3703 & 1000 & 128 & 767 & 745 & 6805 & 8415 \\
{\bf \# Classes} & 7 & 6 & 2 & 18 & 10 & 8 & 10 & 8 \\
{\bf \# Training Nodes} & 140 & 120 & 312 & 4574 & 9625 & 5354 & 12832 & 24144 \\
{\bf \# Validation Nodes} & 300 & 500 & 105 & 1525 & 1376 & 765 & 1834 & 3450 \\
{\bf \# Test Nodes} & 1000 & 1000 & 627 & 1525 & 2751 & 1531 & 3667 & 6899 \\
\bottomrule
\end{tabular}
\end{center}
\end{table*}

\section{Proof of Theorem \ref{thm:injectivity}}
\label{app:injectivity}

Our GOAT layer is a functional composition of the Ordering Part and the Sequence Modeling Part described in Section \ref{sec:GOAT_architecture}. Since the composition of two injective functions is injective itself, it suffices to show that each of the two components is injective. 

We begin by considering the Ordering Part, which maps a multiset of hidden states to an ordered multiset of hidden states leaving the elements of these multisets unchanged. If therefore, for two multisets the same output is generated in the Ordering Part, then their elements are equal. Two multisets with all equal elements are equal themselves. Therefore, the Ordering Part of our GOAT layer is an injective function.

We now consider the Sequence Modeling Part. Following \cite[p.~3]{Hammer2000} we refer to an activation function $\sigma$ as a \textit{squashing activation function} if $\sigma:\mathbb{R}\rightarrow[0,1]$ is monotonous with $\lim_{x\rightarrow-\infty}\sigma(x)=0$ and $\lim_{x\rightarrow\infty}\sigma(x)=1.$ In the now following Theorem \ref{thm:Hammer} we combine and rephrase the universal approximation results in Theorem 3 and Corollary 4 in \cite[pp.~6,~8]{Hammer2000}.

\begin{theorem}\label{thm:Hammer}
Any measurable function $f:\mathcal{X}\rightarrow \mathbb{R}^t$ can be approximated arbitrarily well in probability by a recurrent neural network $\feedfn \circ \tilde{\recfn}_y,$ where $\recfn:\mathbb{R}^{q+1}\rightarrow\mathbb{R}$ is a feedforward neural network without a hidden layer and either a squashing activation function or a locally Riemann integrable and non polynomial activation function. $h:\mathbb{R}\rightarrow\mathbb{R}^t$ is either a linear mapping or a multilayer network with one hidden layer with squashing or locally Riemann integrable and nonpolynomial activation function and linear outputs.    
\end{theorem}

Theorem \ref{thm:Hammer} applies to recurrent neural networks which are formally defined by \cite[p.~3]{Hammer2000} to be composed of two feedforward neural networks $\recfn:\mathbb{R}^{q+1}\rightarrow\mathbb{R}$ and $\feedfn:\mathbb{R}\rightarrow\mathbb{R}^t.$ The function $\recfn$ is referred to as the ``recursive part'' since it gives rise to the function $$\tilde{\recfn}_y([h_1, \ldots, h_i]) = \begin{cases} y& \mathrm{for ~} i=0;\\\recfn(h_i,\tilde{\recfn}_y([h_1, \ldots, h_{i-1}]))& \mathrm{otherwise}.\end{cases}$$
The function $\feedfn$ is simply referred to as the ``feedforward part'' since it succeeds the recursive part and plays the role of a standard readout function in a deep learning architecture. That is to say, that a recurrent neural network $f$ is defined to equal the composition of $f=\feedfn\circ\tilde{\recfn}_y.$ Now all three of the RNNs we consider as possible aggregators, the standard RNN, LSTM and GRU, satisfy the conditions of this formal definition of recurrent neural networks and therefore,  Theorem \ref{thm:Hammer} can be applied to our three considered RNNs.

Now that we have shown that Theorem \ref{thm:Hammer} applies to all RNNs we consider in the Sequence Modeling Part, we can use it to establish that the function learned in our Sequence Modeling Part can approximate any measurable function arbitrarily well in probability. This notably includes all measurable injective functions, thus providing us with the desired result. 



\begin{table*}[t]
\centering
\caption{Comparison between different attention mechanisms in the GOAT layer.
We report the classification accuracy (± standard deviation) on the Cora, CiteSeer, Disease and “Top-2 pooling” datasets and
MSE (± standard deviation) results on the synthetic datasets “Betweenness Centrality” and “Effective
Size” for two different types of random graphs.}
\label{table:gat_gatv2}
\resizebox{\linewidth}{!}{
\begin{tabular}{lccccccccc}
\toprule
\multirow{2}{*}{\textbf{Method}} & \textbf{Cora} & \textbf{CiteSeer} & \textbf{Disease} &
\multicolumn{1}{c}{\textbf{Top-2 pooling}} &  \multicolumn{2}{c}{\textbf{Betweenness Centrality }} &  \multicolumn{2}{c}{\textbf{Effective Size} } \\ & & & & & (100,0.09) & (1000,0.01) & (100,0.09) & (1000,0.01) \\ 
\midrule
\textbf{GOAT(gat)} & \textbf{84.9} & \textbf{69.5} & \textbf{92.11} $\pm 1.88$ &  \textbf{69.21} $\pm 5.10$ & \textbf{0.0038 $\pm 0.0019$} & \textbf{0.0006} $\pm 0.0002$ & 0.0016 $\pm 0.0008$ & 0.0002 $\pm 0.000082$ \\ 
\textbf{GOAT(gatv2)} & 83.1 & 69.3 &  91.28 $\pm 1.75$ & 67.34 $\pm 5.24$ & \textbf{0.0038} $\pm 0.0022$ &  \textbf{0.0006} $\pm 0.0001$ & \textbf{0.0013} $\pm 0.0008$ & \textbf{0.0001} $\pm 0.000037$   \\
\bottomrule
\end{tabular}
}
\end{table*}

\section{Experimental Details}\label{appendix:datasets}

Included in this supplementary material is our implementation, which is built upon the open source library \textit{PyTorch Geometric} (PyG) under MIT license \citep{https://doi.org/10.48550/arxiv.1903.02428}. The experiments are run on an Intel(R) Xeon(R) CPU E5-1607 v2 @ 3.00GHz processor with $128$GB RAM and a NVIDIA Corporation GP102 TITAN X GPU with $12$GB RAM. 

\textbf{Datasets Details.}
In our experiments we utilize nine well-known node classification benchmarks. We describe them below:
\begin{itemize}
    \item 2 citation network benchmark datasets: Cora,
CiteSeer \citep{Sen_Namata_Bilgic_Getoor_Galligher_Eliassi-Rad_2008}, where nodes represent scientific papers, edges are citations between them, and node labels are academic topics. 
We follow the experimental setup of \citet{kipf2017semisupervised} and use 140 nodes for training, 300 for validation and 1000 for testing. 
We optimize hyperparameters on Cora and use the same hyperparameters for CiteSeer. 
\item 1 disease spreading model: Disease \citep{chami2019hyperbolic}. It simulates the SIR disease spreading model \citep{nla.cat-vn2624541}, where the label of a node indicates if it is infected or not. 
We follow the experimental setup of \citet{chami2019hyperbolic} and use 30/10/60\% for training, validation and test sets and report the average results from 10 different splits.
\item 1 social network: LastFM Asia \citep{feather}.
Nodes are LastFM users from Asian countries and edges are mutual follower relationships between them. The label of each node is the country of the user.
We use 60/20/20\% for training, validation and test sets and report the average results from 10 different splits.
\item 2 co-purchase graphs: Amazon Computers, Amazon Photo \citep{shchur2019pitfalls}. Nodes represent products and edges represent that two products are frequently bought together. The node label indicates the product category. 
We use 70/10/20\% for training, validation and test sets and report the average results from 10 different splits. 
We optimize hyperparameters on Computers and use the same hyperparameters for Photo. 
\item 2 co-authorship graphs: Coauthor CS, Coauthor Physics \citep{shchur2019pitfalls}.
Nodes represent authors that are connected by an edge if they co-authored a paper. Given paper keywords for each author’s papers as node features, the task is to identify the field of study of the authors authors.
We use 70/10/20\% for training, validation and test sets and report the average results from 10 different splits. 
\item 1 large citation network: ogbn-arxiv \citep{ogb}. Each node is an arXiv paper and an edge indicates that one paper cites another one. The task is to predict the 40 subject areas of arXiv CS papers.
We use the public split by publication dates provided by the original paper.
\end{itemize}
We report further summary statistics of these datasets in Table \ref{datasets}.

In our synthetic tasks we predic the betweenness centrality $b(u)$ and the effective size $e(u)$. 

\textbf{The betweenness centrality} $b(u)$ is a measure of centrality of a node $u$ based on shortest paths involving $u.$ It has many applications in network science, as it is a useful metric for analyzing communication dynamics \citep{goh2003betweenness}. It can be computed using the following equation 
\begin{equation*}
b(u) = \sum_{s,t \in V} \dfrac{\sigma(s,t|u)}{\sigma(s,t)},
\end{equation*}
where $\sigma(s, t)$ is the number of distinct shortest paths between vertices $s$ and $t,$ and $\sigma(s, t|u)$ is the number of these shortest paths passing through $u.$ 

\textbf{The effective size} $e(u)$ \citep{article_social_networks} of node $u$ is based on the concept of redundancy and for the case of unweighted and undirected graphs, can be computed as 
\useshortskip
\begin{equation*}
    e(u) = n - \dfrac{2q}{n},
\end{equation*}
where $q$ is the number of ties in the subgraph induced by the node set $\closedneighborhood{u}$ (excluding ties involving $u$) and $n = |\neighborhood{u}|$ is the number of neighbors (excluding the central node).

\textbf{Synthetic Experiments: Prediction of Graph Structural Properties (node regression)}
For the GCN and GraphSage model we transform the input features with a linear layer and then use 2 convolutional layers followed by 1 linear layer. To optimize the hyper-parameters we perform a grid-search on the following values:
$\text{linear } = \{4,8,16,32,64\}$ for the first linear layer, 
$\text{conv1 } = \{4,8,16,32,64\}$ for the first convolutional layer, $\text{conv2 }= \{4,8,16,32\}$ for the second convolutional layer.
For the GAT and GOAT model we optimize the following hyper-parameters:
$\text{nheads }= \{1,4,8\}$ for the number of heads, 
$\text{conv1 }= \{4,8,16,32,64\}$ for the first convolutional layer,
$\text{conv2 } =  \{4,8,16,32,64\}$ for the second convolutional layer.
Also for the GOAT model, we use one GOAT layer and one GCN or GAT layer as the second layer. 
Specifically, for the ``Betweenness Centrality'' and ``Effective Size'' tasks we used GAT as the second layer, and for the ``Top-2 pooling'' task we used GCN.
For the PNA model we optimize the following hyper-parameters:
$\text{linear } = \{4,8,16,32,64\}$ for the first linear layer, 
$\text{conv1 } = \{4,8,16,32,64\}$ for the first convolutional layer, $\text{conv2 }= \{4,8,16,32\}$ for the second convolutional layer,
\text{aggregators} = \{`mean', `min', `max', `std'\} for the aggregators
\text{scalers} = \{`identity', `amplification', `attenuation',`linear'\} for the scalers.
We search for the best model on $(N=100,p=0.09)$ and we use the same models for the other configuration of each task $(N=1000,p=0.1).$

\textbf{Node classification Benchmarks.}
For node classification benchmarks we follow the same model configurations as with node regression above and we just remove the last linear layers from all the models. For the GOAT model, we used GAT as a second layer in Cora, CiteSeer and Disease and GCN as a second layer in LastFM Asia, Computers, Photo, CS and Physics datasets. 

\comment{
\textbf{Best configurations found for Betweenness Centrality task:} \\
\textbf{GraphSage (mean)}: linear : 8, conv1 : 16, conv2 : 16 \\
\textbf{GraphSage (lstm)}: linear : 8, conv1 output: 8, conv2 output: 16 \\
\textbf{GIN}: linear : 4 , conv1 : 16 , conv2 output: 8 \\
\textbf{GCN}: linear : 16, conv1 : 64, conv2 output: 32 \\ 

\textbf{GAT}: nheads:1, linear output:32, conv1 :32, conv2 :8 \\
\textbf{GOAT}: number of heads:1, linear output:8, conv1 output:8, conv2 output:8   \\

\textbf{Best configurations found for Effective Size task:} \\
\textbf{GraphSage (mean)}: linear : 8 , conv1 : 16 , conv2 : 32  \\
\textbf{GraphSage (lstm)}: linear : 4 , conv1 : 64 , conv2 : 16 \\
\textbf{GCN}: linear : 4 , conv1 : 64, conv2 : 32  \\ 
\textbf{GIN}: linear : 4 , conv1 : 64 , conv2 : 8 \\
\textbf{GAT}: nheads:8, linear : 64, conv1 : 64, conv2 :4 \\
\textbf{GOAT}: nheads:4, linear : 64, conv1 : 64, conv2 : 64 \\

\paragraph{Node classification}
\paragraph{Experimental Details}

For \textbf{MLP} we used 3 linear layers \\
linear 1: [4,8,16,32,64]
linear 2: [4,8,16,32,64]
 
For \textbf{GCN},\textbf{GraphSAGE} \textbf{GIN} we used 1 linear layer followed by 2 graph convolutional layers.  \\
Hyperparameters: \\
linear output:[4,8,16,32,64] \\
conv1 output:[4,8,16,32,64]
\\
For \textbf{GAT} we used 2 layers \\
Hyperparameters: \\
linear output: [4,8,16,32,64] \\
number of heads: [1,4,8,16,32,64] \\
pooling: [``cat'',``sum''] \\

For \textbf{GOAT} we used 2 layers \\
Hyperparameters: \\
linear output 1: [4,8,16,32,64] \\
number of heads: [1,4,8,16,32,64] \\
pooling: [``cat'',``sum''] \\

Best hyperparameters found: \\  

\textbf{Cora and we used the same for CiteSeer} \\
MLP: 32,32
GCN: 64,64
GraphSAGE (mean): 32,16
GraphSAGE (lstm): 64,64
GIN: 32,16
GAT: 8,8,cat
GOAT:

\textbf{Disease} \\
MLP: 32,32
GCN: 64,64 
GraphSAGE (mean): 32,16
GraphSAGE (lstm): 8,16
GIN: 16,32
GAT: 8,8,cat
GOAT: 32,1,sum

\textbf{LastFM Asia} \\
MLP: 64,64
GCN : 32,32
GraphSAGE (mean): 32, 32
GraphSAGE (lstm): 16, 8
GIN: 16,64
GAT : 16, 8, cat
GOAT:
\\
\textbf{Amazon and we used the same for Photo}
MLP: 64,8
GCN: 16,16
GraphSAGE (mean): 8,8
GraphSAGE (lstm): 8,8
GIN: 16,8
GAT: 64, 12, cat
GOAT: 8,8,sum
\\
\textbf{Cornell,Texas,Wisconsin} \\
We found the best hyperparameters for Cornell and used the same ones in Texas,Wisconsin
MLP: 8,8
GCN: 8,8
GraphSage (mean): 8, 32
GraphSage (lstm): 8, 32
GIN: 16, 16
GAT: 16, 8, ``cat''
GOAT: 32, 8, ``cat''
}

\section{Additional Experiments} \label{sec:additional_experiments}

\subsection{Comparison of GAT and GATv2} \label{app:gat_vs_gatv2}

We have performed additional experiments investigating the GATv2 attention mechanism as part of a GOAT layer and found it to yield comparable performance to the original GAT attention mechanism.
We follow the same setup as the main experiments and we use an LSTM to aggregate the hidden states of the ordered neighbors.
The results are reported in  Table \ref{table:gat_gatv2}. 

\subsection{Comparison with Janossy Pooling}\label{app:janossy}

We furthermore investigated how our GOAT model compares to the Janossy Pooling model. In Table \ref{results:janossy} we observe the GOAT model to significantly outperform three different hyperparametrizations of the Janossy Pooling model on the Cora, CiteSeet and Disease datasets. 

\begin{table}[t]
\centering
    \caption{Comparison of the accuracies attained by our GOAT architecture and three different hyperparametrizations, given in the format $(k_1,k_2),$ of the Janossy Pooling model\citep{murphy2019janossy}.}
    \label{results:janossy}
    \small
    \begin{tabular}{lcccc}
    \toprule
    \textbf{Method} & \textbf{Cora}  & \textbf{CiteSeer} & \textbf{Disease} & \\
    \midrule
    Janossy Pooling(5,5) & 79.0 & 64.2 & 87.21 $\pm 1.93$ \\
    Janossy Pooling(15,5) & 80.8 & 65.8 & 87.15 $\pm 1.86$ \\
    Janossy Pooling(20,20) & 80.2 & 64.7 & 87.19 $\pm 1.94$\\
    \hline
    \textbf{GOAT(lstm)} & \textbf{84.9} & \underline{69.5} & \textbf{92.11} $\pm 1.88$ \\
    \textbf{GOAT(gru)} & 83.5 &  \textbf{70.0} & \underline{91.97} $\pm 1.90$ & \\
    \textbf{GOAT(rnn)} & \underline{84.2} & 67.9 & 91.67 $\pm 1.69$ & \\ 
    \bottomrule
    \end{tabular}
    
\end{table}

\subsection{Varying the Number of Attention Heads in GOAT}\label{app:att_heads}

In this experiment, we examine the performance of our model using different numbers of attention heads. We use the standard configuration of our GOAT model, i.e., a GAT  attention mechanism paired with an LSTM aggregator. We report the results in Table \ref{results:att_heads}.

We observe that for datasets, for which the learning tasks are known to be relatively simple, i.e., the Cora dataset, a small number of attention heads is sufficient to achieve the best performance. For datasets with complex node interactions and high amount of synergistic information, learning a large number of neighborhood orderings, appears to be beneficial. 

\begin{table}[t]
\centering
    \caption{Accuracy scores of the GOAT architecture when the number of attention heads is varied.}
    \label{results:att_heads}
    \small
    \begin{tabular}{lcccc}
    \toprule
    \textbf{Method} & \textbf{Cora}  & \textbf{CiteSeer} & \textbf{Disease} & \\
    \midrule
    \textbf{GOAT(lstm)$_{1h}$} & \textbf{84.9} & 67.9 & 89.14 $\pm 2.99$ \\
    \textbf{GOAT(lstm)$_{2h}$} & 83.1 & 68.2 & 90.78 $\pm 1.93$\\
    \textbf{GOAT(lstm)$_{4h}$} & 82.8 & \textbf{69.5} & 91.32 $\pm 2.71$\\
    \textbf{GOAT(lstm)$_{8h}$} & 82.9 & 68.8 & \textbf{92.11} $\pm 1.88$ \\
    \bottomrule
    \end{tabular}
\end{table}
\end{document}

%% file: math_commands.tex
\usepackage{amsmath,amsfonts,bm}

\def\eqref#1{equation~\ref{#1}}

\def\1{\bm{1}}

\def\mA{{\bm{A}}}

\def\mD{{\bm{D}}}

\def\mH{{\bm{H}}}
\def\mI{{\bm{I}}}

\def\mS{{\bm{S}}}

\def\mX{{\bm{X}}}

\DeclareMathAlphabet{\mathsfit}{\encodingdefault}{\sfdefault}{m}{sl}
\SetMathAlphabet{\mathsfit}{bold}{\encodingdefault}{\sfdefault}{bx}{n}

\def\gN{{\mathcal{N}}}

\newcommand{\R}{\mathbb{R}}



%% file: goat.bbl
\begin{thebibliography}{40}
\providecommand{\natexlab}[1]{#1}
\providecommand{\url}[1]{\texttt{#1}}
\expandafter\ifx\csname urlstyle\endcsname\relax
  \providecommand{\doi}[1]{doi: #1}\else
  \providecommand{\doi}{doi: \begingroup \urlstyle{rm}\Url}\fi

\bibitem[Scarselli et~al.(2009)Scarselli, Gori, Tsoi, Hagenbuchner, and
  Monfardini]{4700287}
Franco Scarselli, Marco Gori, Ah~Chung Tsoi, Markus Hagenbuchner, and Gabriele
  Monfardini.
\newblock The graph neural network model.
\newblock \emph{IEEE Trans. Neural Netw.}, 20\penalty0 (1):\penalty0 61--80,
  2009.
\newblock \doi{10.1109/TNN.2008.2005605}.

\bibitem[Kipf and Welling(2017)]{kipf2017semisupervised}
Thomas~N. Kipf and Max Welling.
\newblock Semi-supervised classification with graph convolutional networks.
\newblock In \emph{ICLR}, 2017.

\bibitem[Bronstein et~al.(2021)Bronstein, Bruna, Cohen, and
  Veličković]{bronstein2021geometric}
Michael~M. Bronstein, Joan Bruna, Taco Cohen, and Petar Veličković.
\newblock Geometric deep learning: Grids, groups, graphs, geodesics, and
  gauges.
\newblock \emph{arXiv:2104.13478}, 2021.

\bibitem[Gilmer et~al.(2017)Gilmer, Schoenholz, Riley, Vinyals, and
  Dahl]{gilmer2017neural}
Justin Gilmer, Samuel~S Schoenholz, Patrick~F Riley, Oriol Vinyals, and
  George~E Dahl.
\newblock Neural message passing for quantum chemistry.
\newblock In \emph{ICML}, pages 1263--1272, 2017.

\bibitem[Xu et~al.(2019)Xu, Hu, Leskovec, and Jegelka]{xu2019powerful}
Keyulu Xu, Weihua Hu, Jure Leskovec, and Stefanie Jegelka.
\newblock How powerful are graph neural networks?
\newblock In \emph{ICLR}, 2019.

\bibitem[Pei et~al.(2020)Pei, Wei, Chang, Lei, and Yang]{pei2020geomgcn}
Hongbin Pei, Bingzhe Wei, Kevin Chen-Chuan Chang, Yu~Lei, and Bo~Yang.
\newblock Geom-gcn: Geometric graph convolutional networks.
\newblock In \emph{ICLR}, 2020.

\bibitem[Murphy et~al.(2019)Murphy, Srinivasan, Rao, and
  Ribeiro]{murphy2019janossy}
Ryan~L. Murphy, Balasubramaniam Srinivasan, Vinayak Rao, and Bruno Ribeiro.
\newblock Janossy pooling: Learning deep permutation-invariant functions for
  variable-size inputs.
\newblock In \emph{ICLR}, 2019.

\bibitem[Wagstaff et~al.(2021)Wagstaff, Fuchs, Engelcke, Osborne, and
  Posner]{wagstaff2021universal}
Edward Wagstaff, Fabian~B. Fuchs, Martin Engelcke, Michael~A. Osborne, and
  Ingmar Posner.
\newblock Universal approximation of functions on sets.
\newblock \emph{arXiv:2107.01959}, 2021.

\bibitem[Bizzi and Cheung(2013)]{10.3389/fncom.2013.00051}
Emilio Bizzi and Vincent C.~K. Cheung.
\newblock The neural origin of muscle synergies.
\newblock \emph{Front. Comput. Neurosci}, 7:\penalty0 51, 2013.
\newblock ISSN 1662-5188.
\newblock \doi{10.3389/fncom.2013.00051}.
\newblock URL
  \url{https://www.frontiersin.org/article/10.3389/fncom.2013.00051}.

\bibitem[Brenner et~al.(2000)Brenner, Strong, Koberle, Bialek, and
  Steveninck]{article_synergy}
Naama Brenner, Steven Strong, Roland Koberle, William Bialek, and R~Steveninck.
\newblock Synergy in a neural code.
\newblock \emph{Neural computation}, 12:\penalty0 1531--52, 08 2000.
\newblock \doi{10.1162/089976600300015259}.

\bibitem[Pérez-Pérez et~al.(2009)Pérez-Pérez, Candela, and
  Micol]{PEREZPEREZ2009368}
José~Manuel Pérez-Pérez, Héctor Candela, and José~Luis Micol.
\newblock Understanding synergy in genetic interactions.
\newblock \emph{Trends in Genetics}, 25\penalty0 (8):\penalty0 368--376, 2009.
\newblock ISSN 0168-9525.
\newblock \doi{https://doi.org/10.1016/j.tig.2009.06.004}.
\newblock URL
  \url{https://www.sciencedirect.com/science/article/pii/S0168952509001322}.

\bibitem[Williams and Beer(2010)]{pid}
Paul~L. Williams and Randall~D. Beer.
\newblock Nonnegative decomposition of multivariate information.
\newblock \emph{arXiv:1004.2515}, 2010.
\newblock URL \url{http://arxiv.org/abs/1004.2515}.

\bibitem[Vinyals et~al.(2016)Vinyals, Bengio, and
  Kudlur]{DBLP:journals/corr/VinyalsBK15}
Oriol Vinyals, Samy Bengio, and Manjunath Kudlur.
\newblock Order matters: Sequence to sequence for sets.
\newblock In \emph{ICLR}, 2016.
\newblock URL \url{http://arxiv.org/abs/1511.06391}.

\bibitem[Hamilton et~al.(2017)Hamilton, Ying, and
  Leskovec]{hamilton2018inductive}
William~L. Hamilton, Rex Ying, and Jure Leskovec.
\newblock Inductive representation learning on large graphs.
\newblock In \emph{NIPS}, 2017.

\bibitem[Gori et~al.(2005)Gori, Monfardini, and Scarselli]{1555942}
M.~Gori, G.~Monfardini, and F.~Scarselli.
\newblock A new model for learning in graph domains.
\newblock In \emph{IJCNN}, pages 729--734, 2005.
\newblock \doi{10.1109/IJCNN.2005.1555942}.

\bibitem[Wagstaff et~al.(2019)Wagstaff, Fuchs, Engelcke, Posner, and
  Osborne]{wagstaff2019limitations}
Edward Wagstaff, Fabian~B. Fuchs, Martin Engelcke, Ingmar Posner, and Michael
  Osborne.
\newblock On the limitations of representing functions on sets.
\newblock In \emph{ICML}, 2019.

\bibitem[Veličković et~al.(2018{\natexlab{a}})Veličković, Cucurull,
  Casanova, Romero, Liò, and Bengio]{velikovi2017graph}
Petar Veličković, Guillem Cucurull, Arantxa Casanova, Adriana Romero, Pietro
  Liò, and Yoshua Bengio.
\newblock Graph attention networks.
\newblock In \emph{ICLR}, 2018{\natexlab{a}}.

\bibitem[Brody et~al.(2022)Brody, Alon, and Yahav]{brody2022how}
Shaked Brody, Uri Alon, and Eran Yahav.
\newblock How attentive are graph attention networks?
\newblock In \emph{International Conference on Learning Representations}, 2022.
\newblock URL \url{https://openreview.net/forum?id=F72ximsx7C1}.

\bibitem[Zhang and Xie(2020)]{Zhang_2020}
Shuo Zhang and Lei Xie.
\newblock Improving attention mechanism in graph neural networks via
  cardinality preservation.
\newblock In \emph{IJCAI}, 2020.
\newblock ISBN 9780999241165.
\newblock \doi{10.24963/ijcai.2020/194}.
\newblock URL \url{http://dx.doi.org/10.24963/ijcai.2020/194}.

\bibitem[Niepert et~al.(2016)Niepert, Ahmed, and Kutzkov]{Niepert2016}
Mathias Niepert, Mohamed Ahmed, and Konstantin Kutzkov.
\newblock Learning convolutional neural networks for graphs.
\newblock In \emph{International conference on machine learning (ICML)}, pages
  2014--2023. PMLR, 2016.

\bibitem[Gao et~al.(2018)Gao, Wang, and Ji]{Gao2018}
Hongyang Gao, Zhengyang Wang, and Shuiwang Ji.
\newblock Large-scale learnable graph convolutional networks.
\newblock In \emph{Proceedings of the 24th ACM SIGKDD international conference
  on knowledge discovery \& data mining}, pages 1416--1424, 2018.

\bibitem[Veličković et~al.(2018{\natexlab{b}})Veličković, Fedus, Hamilton,
  Liò, Bengio, and Hjelm]{dgi}
Petar Veličković, William Fedus, William~L. Hamilton, Pietro Liò, Yoshua
  Bengio, and R~Devon Hjelm.
\newblock Deep graph infomax.
\newblock In \emph{ICLR}, 2018{\natexlab{b}}.

\bibitem[Peng et~al.(2020)Peng, Huang, Luo, Zheng, Rong, Xu, and
  Huang]{mi_survey}
Zhen Peng, Wenbing Huang, Minnan Luo, Qinghua Zheng, Yu~Rong, Tingyang Xu, and
  Junzhou Huang.
\newblock Graph representation learning via graphical mutual information
  maximization.
\newblock In \emph{WWW}, 2020.
\newblock URL \url{https://arxiv.org/abs/2002.01169}.

\bibitem[Luo et~al.(2021)Luo, Li, Peng, Yang, Sun, Yu, and He]{graphentropy}
Gongxu Luo, Jianxin Li, Hao Peng, Carl Yang, Lichao Sun, Philip~S. Yu, and
  Lifang He.
\newblock Graph entropy guided node embedding dimension selection for graph
  neural networks.
\newblock \emph{arXiv:2105.03178}, 2021.
\newblock URL \url{https://arxiv.org/abs/2105.03178}.

\bibitem[Dasoulas et~al.(2020)Dasoulas, Nikolentzos, Scaman, Virmaux, and
  Vazirgiannis]{vnestruct}
George Dasoulas, Giannis Nikolentzos, Kevin Scaman, Aladin Virmaux, and
  Michalis Vazirgiannis.
\newblock Ego-based entropy measures for structural representations.
\newblock \emph{ICASSP}, 2020.
\newblock URL \url{https://arxiv.org/abs/2003.00553}.

\bibitem[Sen et~al.(2008)Sen, Namata, Bilgic, Getoor, Galligher, and
  Eliassi-Rad]{Sen_Namata_Bilgic_Getoor_Galligher_Eliassi-Rad_2008}
Prithviraj Sen, Galileo Namata, Mustafa Bilgic, Lise Getoor, Brian Galligher,
  and Tina Eliassi-Rad.
\newblock Collective classification in network data.
\newblock \emph{AI Mag.}, 29\penalty0 (3):\penalty0 93, 2008.
\newblock \doi{10.1609/aimag.v29i3.2157}.
\newblock URL
  \url{https://ojs.aaai.org/index.php/aimagazine/article/view/2157}.

\bibitem[Hochreiter and Schmidhuber(1997)]{10.1162/neco.1997.9.8.1735}
Sepp Hochreiter and Jürgen Schmidhuber.
\newblock {Long Short-Term Memory}.
\newblock \emph{Neural Comput.}, 9\penalty0 (8):\penalty0 1735--1780, 11 1997.
\newblock ISSN 0899-7667.
\newblock \doi{10.1162/neco.1997.9.8.1735}.
\newblock URL \url{https://doi.org/10.1162/neco.1997.9.8.1735}.

\bibitem[Hammer(2000)]{Hammer2000}
Barbara Hammer.
\newblock On the approximation capability of recurrent neural networks.
\newblock \emph{Neurocomputing}, 31\penalty0 (1-4):\penalty0 107--123, 2000.

\bibitem[Corso et~al.(2020)Corso, Cavalleri, Beaini, Li\`{o}, and
  Veli\v{c}kovi\'{c}]{corso2020pna}
Gabriele Corso, Luca Cavalleri, Dominique Beaini, Pietro Li\`{o}, and Petar
  Veli\v{c}kovi\'{c}.
\newblock Principal neighbourhood aggregation for graph nets.
\newblock In \emph{Advances in Neural Information Processing Systems}, 2020.

\bibitem[Cho et~al.(2014)Cho, van Merri{\"e}nboer, Bahdanau, and
  Bengio]{cho-etal-2014-properties}
Kyunghyun Cho, Bart van Merri{\"e}nboer, Dzmitry Bahdanau, and Yoshua Bengio.
\newblock On the properties of neural machine translation: Encoder{--}decoder
  approaches.
\newblock In \emph{Proceedings of {SSST}-8, Eighth Workshop on Syntax,
  Semantics and Structure in Statistical Translation}, pages 103--111, 2014.

\bibitem[Kingma and Ba(2015)]{kingma2017adam}
Diederik~P. Kingma and Jimmy Ba.
\newblock Adam: A method for stochastic optimization.
\newblock In \emph{ICLR}, 2015.

\bibitem[Hu et~al.(2020)Hu, Fey, Zitnik, Dong, Ren, Liu, Catasta, and
  Leskovec]{ogb}
Weihua Hu, Matthias Fey, Marinka Zitnik, Yuxiao Dong, Hongyu Ren, Bowen Liu,
  Michele Catasta, and Jure Leskovec.
\newblock Open graph benchmark: Datasets for machine learning on graphs.
\newblock \emph{arXiv:2005.00687}, 2020.

\bibitem[Chami et~al.(2019)Chami, Ying, Ré, and Leskovec]{chami2019hyperbolic}
Ines Chami, Rex Ying, Christopher Ré, and Jure Leskovec.
\newblock Hyperbolic graph convolutional neural networks.
\newblock \emph{NeurIPS}, 2019.

\bibitem[Rozemberczki and Sarkar(2020)]{feather}
Benedek Rozemberczki and Rik Sarkar.
\newblock {Characteristic Functions on Graphs: Birds of a Feather, from
  Statistical Descriptors to Parametric Models}.
\newblock In \emph{CIKM}, page 1325–1334. ACM, 2020.

\bibitem[Shchur et~al.(2019)Shchur, Mumme, Bojchevski, and
  Günnemann]{shchur2019pitfalls}
Oleksandr Shchur, Maximilian Mumme, Aleksandar Bojchevski, and Stephan
  Günnemann.
\newblock Pitfalls of graph neural network evaluation.
\newblock In \emph{R2L Workshop at NeurIPS}, 2019.

\bibitem[Kim and Oh(2022)]{https://doi.org/10.48550/arxiv.2204.04879}
Dongkwan Kim and Alice Oh.
\newblock How to find your friendly neighborhood: Graph attention design with
  self-supervision, 2022.
\newblock URL \url{https://arxiv.org/abs/2204.04879}.

\bibitem[Fey and Lenssen(2019)]{https://doi.org/10.48550/arxiv.1903.02428}
Matthias Fey and Jan~Eric Lenssen.
\newblock Fast graph representation learning with pytorch geometric, 2019.
\newblock URL \url{https://arxiv.org/abs/1903.02428}.

\bibitem[Anderson and May(1992)]{nla.cat-vn2624541}
Roy~M Anderson and Robert~M May.
\newblock \emph{Infectious diseases of humans: dynamics and control}.
\newblock Oxford university press, 1992.

\bibitem[Goh et~al.(2003)Goh, Oh, Kahng, and Kim]{goh2003betweenness}
K-I Goh, Eulsik Oh, Byungnam Kahng, and Doochul Kim.
\newblock Betweenness centrality correlation in social networks.
\newblock \emph{Phys. Rev. E}, 67\penalty0 (1):\penalty0 017101, 2003.

\bibitem[Everett and Borgatti(2020)]{article_social_networks}
Martin Everett and Stephen Borgatti.
\newblock Unpacking burt’s constraint measure.
\newblock \emph{Soc. Netw.}, 62:\penalty0 50--57, 07 2020.
\newblock \doi{10.1016/j.socnet.2020.02.001}.

\end{thebibliography}
